\pdfoutput=1

\documentclass[11pt]{article}

\usepackage{EMNLP2022}

\usepackage{times}
\usepackage{latexsym}

\usepackage[T1]{fontenc}

\usepackage[utf8]{inputenc}

\usepackage{microtype}

\usepackage{inconsolata}

\usepackage{graphicx}
\usepackage{subfigure}
\usepackage{xfrac}
\usepackage{colortbl}
\usepackage{booktabs}  
\usepackage{amsfonts}  
\usepackage{nicefrac}
\usepackage{latexsym}
\usepackage{multirow}
\usepackage{amsmath}
\usepackage{tabularx}
\usepackage{todonotes} 

\definecolor{gred}{RGB}{219,68,55}
\definecolor{gblue}{RGB}{66,133,244}
\definecolor{gyellow}{RGB}{244,180,0}
\definecolor{ggreen}{RGB}{85,157,88}
\definecolor{ggrey}{RGB}{115,115,115}
\definecolor{na}{gray}{0.9}

\newcommand{\colorR}[1]{\textcolor{gred}{#1}}
\newcommand{\colorG}[1]{\textcolor{ggreen}{#1}}
\newcommand{\colorB}[1]{\textcolor{gblue}{#1}}
\newcommand{\colorY}[1]{\textcolor{gyellow}{#1}}

\newcommand*{\affmark}[1][*]{\textsuperscript{#1}}
\newcommand*{\email}[1]{\texttt{#1}}

\usepackage{booktabs}
\newcommand{\tabitem}{~~\quad\llap{\textbullet}~~}
\newcommand{\tabsubitem}{~~\quad\quad\llap{\textbullet}~~}

\title{Unsupervised Multi-Granularity Summarization}

\author{
Ming Zhong\affmark[$\S$]\thanks{Ming completed this work during his internship at Microsoft.}
\bf \quad Yang Liu\affmark[$\dagger$]
\quad Suyu Ge\affmark[$\S$]
\quad Yuning Mao\affmark[$\S$]
\quad Yizhu Jiao\affmark[$\S$]\\
\bf \quad Xingxing Zhang\affmark[$\ddagger$]
\bf \quad Yichong Xu\affmark[$\dagger$]
\bf \quad Chenguang Zhu\affmark[$\dagger$]
\bf \quad Michael Zeng\affmark[$\dagger$]
\bf \quad Jiawei Han\affmark[$\S$]
\\
{\affmark[$\S$]University of Illinois at Urbana-Champaign} \\
{\affmark[$\dagger$]Microsoft Cognitive Services Research} \quad
{\affmark[$\ddagger$]Microsoft Research Asia}  \\
\email{\{mingz5, suyuge2, yuningm2, yizhuj2, hanj\}}@illinois.edu\\
\email{\{yaliu10, xizhang, yichong.xu, chezhu, nzeng\}}@microsoft.com}

\begin{document}
\maketitle
\begin{abstract}
Text summarization is a user-preference based task, i.e., for one document, users often have different priorities for summary.
As a key aspect of customization in summarization, granularity is used to measure the semantic coverage between summary and source document.
However, developing systems that can generate summaries with customizable semantic coverage is still an under-explored topic.
In this paper, we propose the first unsupervised multi-granularity summarization framework, \textsc{GranuSum}.
We take events as the basic semantic units of the source documents and propose to rank these events by their salience.
We also develop a model to summarize input documents with given events as anchors and hints.
By inputting different numbers of events, \textsc{GranuSum} is capable of producing multi-granular summaries in an unsupervised manner.
Meanwhile, we annotate a new benchmark \textit{GranuDUC} that contains multiple summaries at different granularities for each document cluster. 
Experimental results confirm the substantial superiority of  \textsc{GranuSum} on multi-granularity summarization over strong baselines. 
Further, by exploiting the event information, \textsc{GranuSum} also exhibits state-of-the-art performance under conventional unsupervised abstractive setting.\footnote{Dataset for this paper can be found at: \url{https://github.com/maszhongming/GranuDUC}.}
\end{abstract}

\section{Introduction}
Text summarization aims to condense and summarize long documents into a concise paragraph containing the essential points of the original texts~\cite{see2017get,liu2019text,wang2020heterogeneous,DBLP:conf/acl/ZhongLCWQH20,liu2022brio,an2022colo}.
Notably, the requirements for summarization are highly customized and personalized for different users~\cite{diaz2007user,lerman2009sentiment,yan2011summarize,chen2021dialogsum}. Therefore, generating quality summaries to meet different preferences should be a natural capability of summarization systems.

\renewcommand\arraystretch{0.6}
\begin{table}[t]
    \centering \footnotesize
    \tabcolsep0.01 in
    \begin{tabular}{p{3in}}
    \toprule
    \textbf{Multiple News Articles about Hurricane Mitch} \\
    \midrule
    Honduras braced for potential catastrophe Tuesday as Hurricane Mitch roared through the northwest Caribbean, churning up high waves and intense rain ... (\textbf{Total 3,358 words}) \\
    \midrule
    \textbf{Summary of Coarse Granularity Level} \\
    \midrule
    \colorB{Hurricane Mitch, category 5 hurricane}, brought widespread death and destruction to \colorB{Central American}, and \colorB{Honduras} was especially hard hit. (\textbf{Total 19 words})\\
    \midrule
    \textbf{Summary of Medium Granularity Level} \\
    \midrule
     \colorB{Hurricane Mitch approached Honduras} on Oct. 27, 1998 with winds up to 180mph \colorB{a Category 5 storm} ... \colorR{The European Union, international relief agencies, Mexico, the U.S., Japan, Taiwan, the U.K. and U.N. sent financial aid, relief workers and supplies}. (\textbf{Total 53 words}) \\
    \midrule
    \textbf{Summary of Fine Granularity Level} \\
    \midrule
    \colorB{A category 5 storm, Hurricane Mitch} roared across the northwest Caribbean with 180 mph winds across a 350-mile front ...
    The greatest losses were in \colorB{Honduras} where 6,076 people perished ... At least 569,000 people were homeless across \colorB{Central America}. \colorR{Aid was sent from many sources (European Union, the UN, US and Mexico). The U.S. and European Union were joined by Pope John Paul II in a call for money and workers to help the stricken area}. However, Relief efforts are hampered by extensive damage ... (\textbf{Total 133 words}) \\
    \bottomrule
    \end{tabular}
    \caption{An example from our multi-granularity summarization benchmark GranuDUC. Texts of the same color (blue, red) denote similar points described in different ways. Finer-grained summaries have higher semantic coverage with the original text.}
    \label{tab:intro_case}
    \vspace{-4mm}
\end{table}

Granularity, a key aspect of customization in summarization, is used to measure the degree of semantic coverage between summary and source documents~\cite{mulkar2011granularity}.
To cater to the diverse needs of readers, the granularity level of summaries often varies in a wide range.
As shown in Table~\ref{tab:intro_case}, given multiple news about Hurricane Mitch, the most compact summary (Coarse Granularity Level) accommodates only the most important event to help people grasp the overall picture of the input documents.
Interested readers, on the other hand, may prefer more fine-grained summaries (Medium and Fine Granularity Level) to acquire additional details, such as how many casualties were caused and how different countries aided Honduras.
Thus, multi-granularity summaries can meet the intent of different users and are more versatile in real-world applications. 

Most existing summarization models and benchmarks focus solely on single-granularity summarization.
It limits the ability of these systems to adapt to different user preferences and generalize to a wider range of granularity scenarios. 
To alleviate this issue, some recent studies are dedicated to controlling the length of summary~\cite{kikuchi2016controlling, fan2018controllable, liu2018controlling}.
However, as a surface-level feature of the summary, longer length does not equate to a higher degree of semantic coverage.
In other words, the length limit can be easily satisfied by talking less/more details about the same event, but this is in contrast with the concept of summarization.
Another research direction is query/aspect-based ~\cite{zhong2021qmsum, hayashi2021wikiasp, ge2021fine} and interactive summarization~\cite{shapira2017interactive, shapira2021extending}. Based on different queries, models can focus on different parts of the document and create summaries of various granularities. In practice, it requires a user to provide a query, implying that the user must have prior knowledge of the topic of the source text.
Therefore, automatic granularity-aware summarization model is still an under-explored topic.

In this paper, we propose an unsupervised multi-granularity summarization framework called \textsc{GranuSum}.
Unlike previous work based on supervised learning to provide guidance signals, such as salient sentences~\cite{dou2021gsum}, keywords~\cite{he2020ctrlsum}, and retrieved summaries~\cite{an2021retrievalsum}, our approach does not rely on any manually labeled data.
To measure the granularity, we first regard events as the basic semantic units of the input texts because events carry rich semantic information and are considered as informative representations in many NLP tasks~\cite{zhang2020aser,li2020connecting,chen2021event}.
Overall, our system consists of two event-related components: Event-aware Summarizer and Event Selector. Specifically, given the document and randomly selected events in it as hints, we pre-train an abstractive Summarizer that can recover event-related passages.
Furthermore, in an unsupervised manner, our Event Selector selects the events with high salience from the original text by candidate events pruning and ranking.
Finally, through selecting different numbers of anchor events based on Event Selector, we can control the Summarizer to generate summaries containing different events, thus covering different numbers of semantic units of the original text.
With our proposed approach, \textsc{GranuSum} becomes an unsupervised framework for multi-granularity summary generation.

To evaluate the multi-granularity summarization systems,
we re-annotate DUC2004~\cite{dang2005overview} as the first benchmark in this direction (denoted as \textit{GranuDUC}).
Given multiple documents on the same topic, we annotate summaries at three levels of granularity with different semantic coverage.
Also, to utilize the existing datasets for a supplement evaluation,  we propose to divide several large-scale summarization datasets into buckets with summaries at different granularity levels to further evaluate the model performance.
Experimentally, \textsc{GranuSum} surpasses strong summarization systems on all the multi-granularity evaluations. Additionally, we conduct conventional unsupervised abstractive summarization experiments on three typical benchmarks in different domains. Results demonstrate that \textsc{GranuSum} also substantially improves the previous state-of-the-art model under the traditional setting.

\section{Related Work}

\subsection{Customized Summarization}

In order to meet the needs of different users, existing neural summarization systems attempt to control customization of the summary, such as the aspects of content~\cite{zhong2021qmsum,hayashi2021wikiasp}, summary length~\cite{christensen2014hierarchical,kikuchi2016controlling,liu2018controlling} and writing style~\cite{an2021retrievalsum}. Also, several studies seek to accommodate multiple types of preferences simultaneously to achieve customized summarization. \citet{fan2018controllable} additionally introduces different special marker tokens to the model to generate user-controllable summaries. \citet{he2020ctrlsum} allows for entity-centric, length-controllable, and question-guided summarization by adjusting the prompts, i.e., changing the textual input in the form of a set of keywords or descriptive prompt words.
However, the unavailability of large-scale data containing customized summaries limits the development of these systems that rely on supervised learning.
Thus, we focus on unsupervised approaches and are committed to solving the granularity aspect, which remains an under-explored direction in customized summarization.

\subsection{Unsupervised Summarization}

In contrast to supervised learning, unsupervised models do not require any human-annotated summaries during training. Unsupervised summarization can also be divided into two branches: extractive methods and abstractive approaches. Most extractive methods rank the sentences and select the highest-ranked ones to form the summary. Specifically, they score sentences based on graph~\cite{erkan2004lexrank,hirao2013single,parveen2015topical}, centrality~\cite{zheng2019sentence,liang2021improving}, point-wise mutual information~\cite{padmakumar2021unsupervised}, or sentence-level self-attention in pre-trained models~\cite{xu2020unsupervised}. Another direction is unsupervised abstractive approaches, and these studies typically employ sequence-to-sequence auto-encoding method~\cite{chu2019meansum} with adversarial training and reinforcement learning~\cite{wang2018learning}. In addition, \citet{yang2020ted} pre-train a Transformer model for unsupervised abstractive summarization by exploiting the lead bias phenomenon~\cite{see2017get,zhong2019searching} in the news domain. In this work, our framework is an unsupervised abstractive framework, and can be further enhanced on top of the extractive method.

\section{Multi-Granularity Framework}

\begin{figure}[t]
    \centering
    \includegraphics[width=1\linewidth]{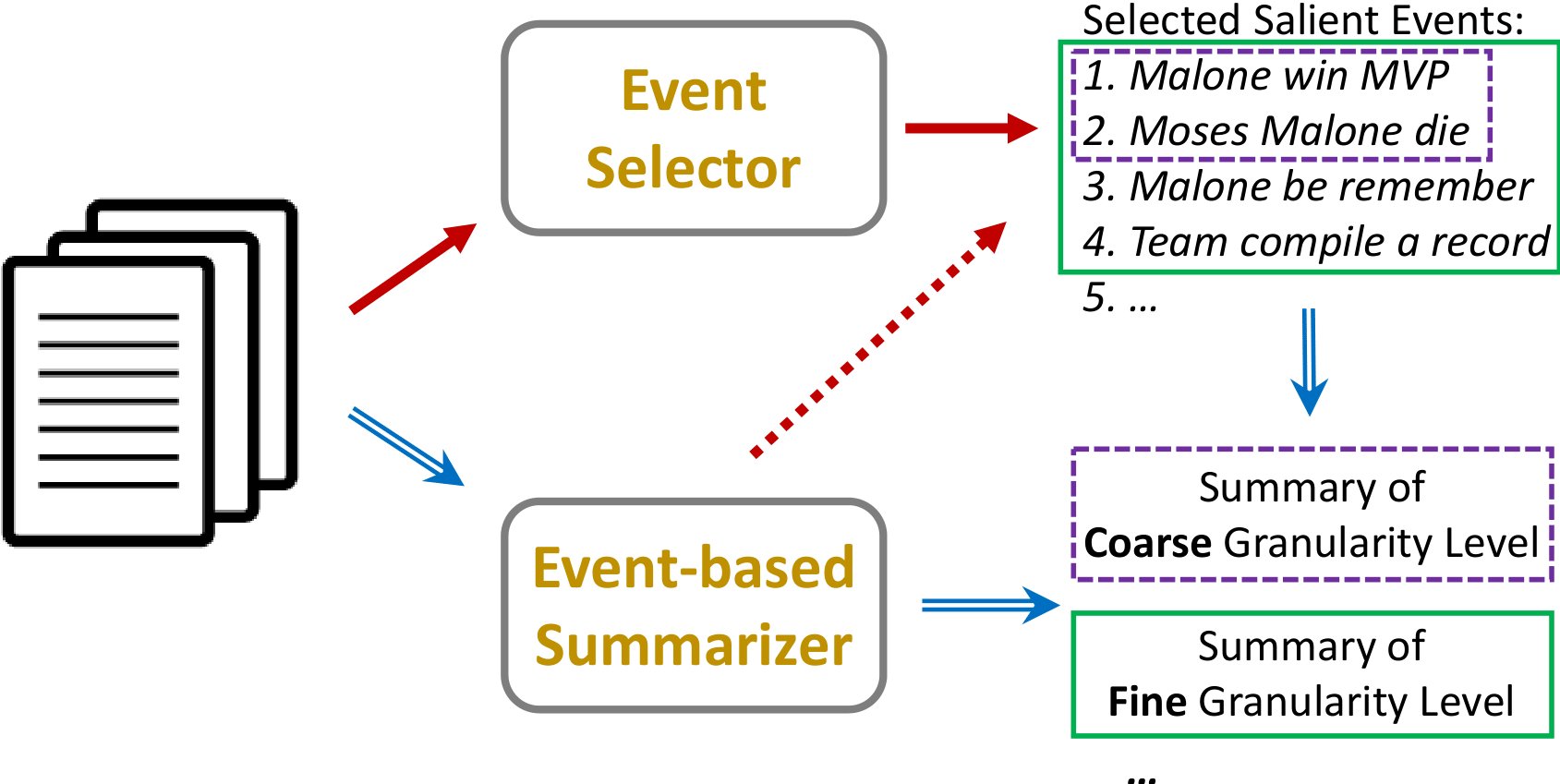}
    \caption{Overview of \textsc{GranuSum}. It consists of two components: Event Selector and Event-aware Summarizer. The red line ($\rightarrow$) indicates that Selector extracts the salient events from the text, and the dotted line means that Summarizer assists in this process. The blue line ($\Rightarrow$) denotes the multi-granularity summary generation process. By inputting different numbers of events as anchors (purple and green boxes), \textsc{GranuSum} can generate multi-granularity summaries.}
    \label{fig:model}
\end{figure}

In this section, we first describe in detail our framework \textsc{GranuSum}, which has two major components: Event-aware Summarizer and Event Selector. Combining them enables multi-granularity generation. The overall framework can be seen in Figure~\ref{fig:model}. Then, we introduce the new human-annotated benchmark, GranuDUC, which can be used for multi-granularity evaluation.

\subsection{Event-Aware Summarizer}
\label{sec:ee}
In this work, we focus on abstractive summarization approaches.
The way we make the model perceive the granularity is by inputting hints with different degrees of specificity, and here we format the hints as a sequence of events.

\paragraph{Event Extraction} We follow previous work to define an event as a verb-centric phrase~\cite{zhang2020aser}.
A lightweight method\footnote{Code for this part is available at: \url{https://github.com/yzjiao/Open-vocabulary-event-extraction}.} is utilized to extract events from open-domain unstructured data: we extract frequently-occurring syntactic patterns that contain verbs as events.
On the basis of \citet{zhang2020aser}, we extend a total of 76 syntactic patterns for matching events.
For instance, the most common patterns contain $n_1$-nsubj-$v_1$ (e.g., \textit{Hurricane hits}) and $n_1$-nsubj-$v_1$-dobj-$n_2$ (e.g., \textit{Earthquake damages buildings}).\footnote{nsubj and dobj indicate nominal subject and direct object. They are different relations between verbs and nouns.}
More details and concrete examples can be found in Appendix~\ref{sec:appendix_ee}.

\paragraph{Event-based Summarizer Pre-training} 
Previous studies reveal that event information can be an effective building block for models to perform text generation~\cite{daniel2003sub,glavavs2014event}, so we attempt to obtain a Summarizer with the ability to generate event-related text in an unsupervised way.
In the pre-training phase, it is trained to regenerate sentences based on a list of events and the remaining source text.
Then we use it to generate a summary at inference time.
Concretely, we pre-train a sequence-to-sequence model in the following steps:

\noindent 1) randomly select a few sentences from the text,

\noindent 2) extract events in these selected sentences,

\noindent 3) mask these sentences in the source document,

\noindent 4) take extracted events and unmasked text as input.

\noindent Then we use these selected sentences as the target for the model.
For example, for a dialogue text as \textit{``Do you have any plans tomorrow? How about playing basketball? Sure, I just finished my homework, it’s time to exercise.''}, we can select \textit{How about playing basketball?} and extract the event \textit{play basketball}.
In this case, the specific format given to the model is:

\begin{itemize}
    \item Input: play basketball $\langle$seg$\rangle$ Do you have any plans tomorrow? $\langle$mask$\rangle$ Sure, I just finished my homework, it's time to exercise.
    \item Target: How about playing basketball?
\end{itemize}

\noindent where $\langle$seg$\rangle$ is the segmentation token and $\langle$mask$\rangle$ indicates that a sentence at this position is masked. 
We use `|` token to split the different events, and another example in news domain to further explain the four steps can be found in Appendix~\ref{sec:appendix_summarizer}.

\subsection{Event Selector}
\label{selector}
The salience of the selected events determines whether the Summarizer can generate a quality summary or an irrelevant and uninformative paragraph.
A long document can contain hundreds of events, and finding the best event subset involves an exponential search space.
Therefore, it is crucial to have an Event Selector that selects the most important events in the text to feed to the Summarizer. Our event selector first reduces the search space by pruning out less salient events and sentences, and then ranks the remaining events using the pre-trained Summarizer.

\paragraph{Event Ranking} The salience of the different events extracted from the documents varies.
Some of the events are informative and relevant to the original text, but others are too general or specific.
For instance, two events \textit{club say} and \textit{Malone be remember} can be extracted from the sentence \textit{``The club said Malone will forever be remembered as a genuine icon and pillar in the Philadelphia 76ers team"}.
The former is not important for this news, while the latter is indispensable.
And in a sentence \textit{``Malone won MVP awards by averaging 24.5 points and 15.3 rebounds"}, \textit{``average 24.5 points and 15.3 rebounds''} is too detailed to be included in a high-level summary.
Thus, ranking candidate events is a key function of Event Selector.

Inspired by \citet{yuan2021bartscore}, where a pre-trained generative model is capable of evaluating the correlation between the input and the target, we also use our pre-trained Event-based Summarizer to calculate the salience score for each event.
Given the candidate event set $E$ and the source document $D$, our Summarizer can generate a candidate summary $c_{E}$.
Whenever an event $e$ in the input is removed, if the generated candidate summary $c_{E\setminus\{e\}}$ differs greatly from $c_{E}$, this indicates that the removed event $e$ is salient.
As in the example above, removing \textit{``club say"} does not cause an obstacle for the model to recover the sentence whose main meaning is that Malone is remembered by people, while removing \textit{``Malone be remember"} makes the model unable to output the correct sentence.
Thus, the latter should be the more important event.
Formally, the \textbf{\underline{Salience Score}} of event $e$ can be defined as:
\begin{align}
    \label{equ:salience}
    \mathrm{Sal}(e) &\stackrel{\text{def}}{=} -\mathrm{Sim}(c_{E\setminus\{e\}};c_{E}), \\
    \mathrm{Sim}(x_1, x_2) &\stackrel{\text{def}}{=}\mathrm{R1(x_1,x_2)+R2(x_1,x_2)},
\end{align}
where $\mathrm{Sim}(x_1, x_2)$ is a function based on ROUGE score~\cite{lin2004rouge} to measure the similarity between any two text sequences $x_1$ and $x_2$. $\mathrm{R1}$ and $\mathrm{R2}$ are ROUGE-1 and ROUGE-2 scores, respectively.
Based on the salience score, Event Selector can rank all the events in the candidate set. However, a single sentence may contain multiple events, so a long document can encompass hundreds of events. Using all events as a candidate set leads to unaffordable computational consumption. Therefore, we prune the candidate events before ranking them.

\paragraph{Candidate Pruning} We expect to capture a small set of events that are relevant to the main topic while pruning redundant parts.
Events with high relevance provide an efficient summary of the central points in the original text, while low redundancy ensures that the final summary is concise.
To this end, we first select several salient sentences and extract the events in them as a candidate set.
For relevance, if a sentence has a high semantic overlap with other input sentences, it should have a higher centrality and a higher probability to be included in the summary~\cite{padmakumar2021unsupervised}. Thus, we define the \textbf{\underline{Relevance Score}} of each sentence as:
\begin{align}
    \mathrm{Rel}(s, D) \stackrel{\text{def}}{=} \mathrm{Sim}(s;D \setminus \{s\}),
\end{align}

\noindent where $s$ means the sentence and $D$ represents the given document. $D\setminus\{s\}$ indicates that the sentence $s$ is removed from the original text $D$.

For redundancy, the sentences in the summary should contain low redundant information when compared with each other.
So when extracting the $k$-th sentence, we define its \textbf{\underline{Redundancy Score}} as follows:
\begin{align}
\mathrm{Red}(s, S) \stackrel{\text{def}}{=} \sum_{i=1}^{k-1} \mathrm{Sim}(s_{i};s),
\end{align}

\noindent where $S$ is a set of the $k$-1 sentences in the summary so far.
We follow the idea of Maximal Marginal Relevance~\cite{DBLP:conf/sigir/CarbonellG98} to maximize relevance and minimize redundancy to calculate the \textbf{\underline{Importance Score}} of each sentence as:
\begin{align}
\mathrm{Imp(s, S, D)} = 
\lambda_{1} \mathrm{Rel}(s, D) - \lambda_{2} \mathrm{Red}(s, S).
\end{align}

Through iteratively calculating the score of each sentence, we can eventually obtain a fixed number of sentences and extract the events from them as a candidate set.

\subsection{Multi-Granularity Summary Generation}
With Event-aware Summarizer and Event Selector, it is feasible to generate multi-granularity summaries.
By taking different numbers of ranked events as hints, the Summarizer can perceive the specific level of semantic coverage required to enable the generation of different summaries.
For example, the Summarizer can generate a concise coarse-grained summary when only the two events with the highest salience scores (see Equation \ref{equ:salience}) are input.
A case study to illustrate the overall flow of the multi-granularity summary generation can be found in Appendix~\ref{sec:appendix_generation}.
During inference, instead of using the same setting as \citet{DBLP:conf/icml/ZhangZSL20}, i.e., placing the $\langle$mask$\rangle$ token at the beginning of the article, we simply omit it.
Because we already provide enough event information to guide the model to generate a summary in our framework.

\subsection{New Benchmark: GranuDUC}
Considering that there is no dataset for evaluating multi-granularity summarization models, we re-annotate a new benchmark called GranuDUC on the basis of DUC2004~\cite{dang2005overview}. Our annotation team consists of 5 graduate students in NLP or people with equivalent expertise. For each document cluster, annotators are required to read multiple source documents and write summaries at three different granularities.
The annotators are informed to be aware that granularity is not distinguished by the number of sentences, but is defined by different semantic coverage of the original text.
Specifically, we inform the annotators that ``coarse granularity level" should include only the main event of the entire documents, ``medium granularity level" should include several important conditions, results and processes surrounding the main topic, and "fine granularity level" should further include the details such as time and location for each sub-event.
Summaries at different granularities require significantly different levels of semantic coverage.
Newly annotated sentences are allowed to be copied or rewritten from DUC2004's original reference summaries. In addition, we require annotators not to use the same sentences in different summaries of a sample, even when describing the same event. Each annotated summary is required to be reviewed by another annotator, then these two people discuss and revise until an agreement is reached. In the end, GranuDUC contains a total of 50 clusters, each cluster contains an average of 10 related documents and 3 summaries of different granularity, ranging from 10 words to more than 200 words in length. To demonstrate the quality of GranuDUC, we include the annotations of two samples in Appendix~\ref{tab:annotation}.

\section{Experiments}
We design three settings of experiments:

\noindent 1) experiments on GranuDUC,

\noindent 2) bucket-based evaluation,

\noindent 3) unsupervised abstractive summarization.

\noindent The first two settings constitute a new testbed for multi-granularity summarization, where bucket means that we divide the existing dataset into different buckets according to semantic coverage to make the evaluation more comprehensive.
In addition to this scenario, the last experiment auxiliarily evaluates the quality of summaries generated by our framework under the conventional unsupervised abstractive summarization setting.

\subsection{Experimental Setup}

\paragraph{Datasets} 
Because the conclusions obtained on the summarization dataset of a single domain are not generalizable~\cite{wang2019exploring, zhong2019closer,chen2020cdevalsumm}, we select two widely varying domains: news and scientific papers for our experiments
Notably, we focus on two types of datasets, multi-document and long-document summarization, which are two main scenarios where users call for a multi-granularity system.
For multi-document summarization, we concatenate the multiple articles into a single sequence as the source text.
In addition to our benchmark GranuDUC, we use the following three datasets. Detailed statistics are listed in Table~\ref{tab:statistics}. 

\underline{Multi-News}~\cite{fabbri2019multi} is a large-scale multi-document summarization dataset in the news domain. We use it in bucket-based evaluation (Section~\ref{bucket_sec}) and unsupervised summarization experiments (Section~\ref{unsupervsied_sec}).

\underline{DUC2004}~\cite{dang2005overview} contains 50 clusters, each with 10 relevant news articles and 4 reference summaries written by humans. Due to its small size, it is usually used directly as a test set. We utilize it in the unsupervised summarization experiment (Section~\ref{unsupervsied_sec}).

\renewcommand\arraystretch{1.1}
\begin{table}
\centering \footnotesize
\tabcolsep0.06 in
\begin{tabular}{lccc}
\toprule
\textbf{Datasets} & \textbf{\# Samples} & \textbf{Len. of Doc.} & \textbf{Len. of Sum.} \\  
\midrule

\textbf{Multi-News} & 56K & 1793 & 217 \\
\textbf{arXiV} & 214K & 6021 & 272 \\
\textbf{DUC2004} & 50 & 5882 & 115 \\

\midrule
\textbf{GranuDUC} & 50 & 5882 & 24/68/135 \\

\bottomrule
\end{tabular}
\caption{Statistics of all datasets we used in this paper. DUC2004 and GranuDUC are for testing only.}
\label{tab:statistics}
\end{table}

\underline{arXiv}~\cite{cohan2018discourse} is a collection of long documents derived from scientific papers. It takes the full text of the paper as input, and the corresponding abstract as the reference summary. We use it in the unsupervised summarization experiment (Section~\ref{unsupervsied_sec}).

\paragraph{Implementation Details} To process long input text in Table~\ref{tab:statistics}, we choose the  Longformer-Encoder-Decoder~(LED)~\cite{beltagy2020longformer} as our backbone model, and train it with typical cross entropy loss.
For Multi-News and arXiv, we further pre-train LED with our event-related generation task on their training corpora  (without using reference summaries) for a total of 10,000 and 30,000 steps, respectively.
We set batch size to 32 and the maximum learning rate to 2e-5. $\lambda_{1}$ in the importance score is 1.0 and $\lambda_{2}$ is 0.4.
By tuning the hyperparameters on the validation set, we empirically extract 9 sentences for Multi-News and 4 sentences for arXiv to form a candidate set, and input 90\% events according to salience score to the Summarizer under unsupervised summarization setting.
For DUC2004 and GranuDUC, we test directly with the Summarizer pre-trained on Multi-News, since these datasets are both in the news domain. In all experiments, we use standard pyrouge\footnote{\url{pypi.python.org/pypi/pyrouge/0.1.3}} to calculate ROUGE scores. Due to the limitation of computational resources, we truncate an input text to 3,072 tokens for LED models.

\paragraph{Baselines}
We use the following baselines:

\underline{BART}~\cite{lewis2020bart} is the state-of-the-art sequence-to-sequence pre-trained model for various generation tasks, including abstractive dialogue generation, question answering, and text summarization. We use BART-large in all the experiments.

\underline{PEGASUS}~\cite{zhang2020pegasus} is a powerful generation model with gap-sentences generation as a pretraining objective tailored for abstractive summarization. We use the large version of PEGASUS for comparison.

\underline{PEGASUS-event} indicates that on top of PEGASUS, additional event information is prepended to the input before the $\langle$mask$\rangle$ token.
We compare it to see if additional event information can be captured without our event-aware pre-training stage.

\underline{LED}~\cite{beltagy2020longformer} has the same architecture as BART, except that the attention in the encoder introduces additional local attention and extends the position embedding to 16K tokens by copying the original embedding. The parameters in the LED are initialized by the weights in BART.

\underline{LED-Length-Control} (LED-LC) is a baseline that we obtained by further pre-training LED. Inspired by \citet{fan2018controllable}, given a document and the desired number of sentences $k$, we randomly place $k$ sentences in the document with the $\langle$mask$\rangle$ token, and let the model recover these sentences. During inference, we input the text and the desired number of sentences as a hint to the model so that it can control the length of the output summary.\footnote{If we need a two-sentence summary, the input format is: ``$\langle$2$\rangle$ $\langle$seg$\rangle$ $\langle$mask$\rangle$ source documents''. It is exactly the same as \textsc{GranuSum} in terms of the training details and data.} 

\underline{PRIMERA}~\cite{xiao2022primera} is a pre-trained model for multi-document summarization that reduces the need for dataset-specific architectures and extensive labeled data. It achieves state-of-the-art results on multi-document summarization datasets under multiple settings.

\subsection{Multi-granularity Evaluation}

The first testbed we built for multi-granularity summarization includes two evaluation methods:

\noindent 1) To test the ability of the model to generate summaries with different granularity levels when given the same input, we evaluate different models on our benchmark GranuDUC.

\noindent 2) To supplement the limited size of GranuDUC, we design a bucket-based evaluation approach, where we divide a large-scale test set into different buckets based on their granularity levels, and test the ability of models to generate quality summaries in different granularity buckets. 

\renewcommand\arraystretch{1.08}
\begin{table*}[t]
\center \footnotesize
\tabcolsep0.11 in
\begin{tabular}{lccccccccc}
\toprule
  &
\multicolumn{3}{c}{\textbf{Coarse Granularity Level}} &
\multicolumn{3}{c}{\textbf{Medium Granularity Level}} &
\multicolumn{3}{c}{\textbf{Fine Granularity Level}} \\

\cmidrule(lr){1-1} \cmidrule(lr){2-4} \cmidrule(lr){5-7} \cmidrule(lr){8-10}
\multicolumn{1}{c}{\textbf{Model}} & \textbf{R-1} & \textbf{R-2} & \textbf{R-L} &
\textbf{R-1} & \textbf{R-2} & \textbf{R-L} &
\textbf{R-1} & \textbf{R-2} & \textbf{R-L} \\

\cmidrule(lr){1-1} \cmidrule(lr){2-4} \cmidrule(lr){5-7} \cmidrule(lr){8-10}

\textsc{PEGASUS} & 20.74 & 4.20 & 15.11 & 24.86 & 4.39 & 14.34 & 29.79 & 5.70 & 14.83 \\
\textsc{PEGASUS}-event & 20.68 & 4.18 & 15.12 & 24.72 & 4.28 & 14.25 & 29.58 & 5.52 & 14.61 \\
\textsc{LED-LC} & 21.83 & 4.80 & 15.29 & 26.73 & 5.59 & 15.76 & 30.18 & 5.57 & 15.24 \\ 

\textsc{GranuSum}  & \textbf{23.61} & \textbf{6.60} & \textbf{17.12} & \textbf{29.69} & \textbf{6.84} & \textbf{16.23} & \textbf{34.71} & \textbf{7.49} & \textbf{17.42} \\ 

\cmidrule(lr){1-1} \cmidrule(lr){2-4} \cmidrule(lr){5-7} \cmidrule(lr){8-10}
\multicolumn{1}{c}{\textbf{Model}} & \textbf{Flu.} & \textbf{Rel.} & \textbf{Faith.} &
\textbf{Flu.} & \textbf{Rel.} & \textbf{Faith.} &
\textbf{Flu.} & \textbf{Rel.} & \textbf{Faith.} \\
\cmidrule(lr){1-1} \cmidrule(lr){2-4} \cmidrule(lr){5-7} \cmidrule(lr){8-10}

\textsc{PEGASUS} & 3.25 & 3.36 & 3.15 & 3.46 & 3.49 & 2.72 & 3.73 & 3.44 & 2.58 \\ 

\textsc{LED-LC} & 3.97 & 3.39 & 3.08 & 3.93 & 3.57 & 3.14 & 3.67 & 3.62 & 2.73 \\ 

\textsc{GranuSum}  & \textbf{4.13} & \textbf{3.82} & \textbf{3.59} & \textbf{4.09} & \textbf{3.78} & \textbf{3.46} & \textbf{3.82} & \textbf{4.05} & \textbf{3.17} \\ 

\bottomrule
\end{tabular}
\caption{Results on GranuDUC. The top half of the Table shows the result of the automatic metric ROUGE, and the bottom half presents the result of human evaluation, including fluency, relevance and faithfulness.}
\label{tab:granuduc}
\end{table*}

\renewcommand\arraystretch{1.1}
\begin{table*}[t]
\center \footnotesize
\tabcolsep0.11 in
\begin{tabular}{lccccccccc}
\toprule
\multicolumn{1}{c}{\multirow{2}[1]{*}{\textbf{Model}}}  &
\multicolumn{3}{c}{\textbf{Low}} &
\multicolumn{3}{c}{\textbf{Medium}} &
\multicolumn{3}{c}{\textbf{High}} \\

 & \textbf{R-1} & \textbf{R-2} & \textbf{R-L} &
\textbf{R-1} & \textbf{R-2} & \textbf{R-L} &
\textbf{R-1} & \textbf{R-2} & \textbf{R-L} \\

\cmidrule(lr){1-1} \cmidrule(lr){2-4} \cmidrule(lr){5-7} \cmidrule(lr){8-10}

\textsc{PRIMERA} & 37.21 & 9.92 & 17.68 & 42.50 & 13.19 & \textbf{20.24} & 46.95 & 18.10 & 23.99 \\

\textsc{LED-LC} & 37.28 & 9.56 & 16.64 & 42.37 & 12.65 & 19.15 & 47.57 & 17.88 & 22.40 \\

\textsc{GranuSum}  & \textbf{38.19} & \textbf{10.27} & \textbf{18.07} & \textbf{44.73} & \textbf{14.12} & 20.10 & \textbf{50.23} & \textbf{19.62} & \textbf{24.11} \\
\quad - Ranking & 37.34 & 9.36 & 16.69 & 43.41 & 13.28 & 19.12 & 49.66 & 19.35 & 23.37 \\

\bottomrule
\end{tabular}
\caption{Result of bucket-based evaluation on Multi-news. We design Granularity Score to divide the test set into three buckets. Low means that the summary has low semantic coverage with the source documents.}
\label{tab:bucket}
\end{table*}

\subsubsection{Results on GranuDUC}
The summaries of each sample in GranuDUC can be divided into three granularity levels, where coarse granularity level represents the most compact summary, and fine granularity level is the most fine-grained summary. We use automatic metrics ROUGE and perform the human evaluation to evaluate the performance of different models in GranuDUC. Notably, both LED-LC and \textsc{GranuSum} have the ability to adjust the output according to specific granularity scenarios. At three different granularity levels on GranuDUC, we let LED-LC output 1, 3 and 8 sentences which correspond to the average length of reference summaries at different granularities. For our model, we take the top 90\% events with the highest salience score in the selected 1, 3, 8 sentences as the input hint.
For all baselines, we control the length of the model output to be similar to the reference summary to get the best performance.

\renewcommand\arraystretch{1.0}
\begin{table*}[t]
\center \footnotesize
\tabcolsep0.15 in
\begin{tabular}{lccccccccc}
\toprule
\multicolumn{1}{c}{\multirow{2}[1]{*}{\textbf{Model}}}  &
\multicolumn{3}{c}{\textbf{Multi-News}} &
\multicolumn{3}{c}{\textbf{arXiv}} &
\multicolumn{3}{c}{\textbf{DUC2004}} \\

 & \textbf{R-1} & \textbf{R-2} & \textbf{R-L} &
\textbf{R-1} & \textbf{R-2} & \textbf{R-L} &
\textbf{R-1} & \textbf{R-2} & \textbf{R-L} \\

\cmidrule(lr){1-1} \cmidrule(lr){2-4} \cmidrule(lr){5-7} \cmidrule(lr){8-10}

LEAD & 42.9 & 14.3 & 19.2 & 32.7 & 8.1 & 17.5 & 32.3 & 6.5 & 16.3 \\

\cmidrule(lr){1-1} \cmidrule(lr){2-4} \cmidrule(lr){5-7} \cmidrule(lr){8-10}

\textsc{LED} & 17.3 & 3.7 & 10.4 & 15.0 & 3.1	& 10.8 & 16.6 & 3.0 & 12.0 \\
\textsc{BART} & 27.3 & 6.2 & 15.1 &  29.2 & 7.5 & 16.9 & 24.1 & 4.0 & 15.3 \\
\textsc{PEGASUS} & 32.0 & 10.1 & 16.7 & 29.5 & 7.9 & 17.1 & 32.7 & 7.4 & 17.6 \\
PEGASUS-event  & 31.5 & 10.2 & 15.8 & 29.2 & 7.7 & 17.0 & 31.8 & 7.1 & 16.9 \\
\textsc{PRIMERA} & 42.2 & 13.7 & \textbf{20.6} & 34.6 & 9.4 & 18.3 & 34.7 & 6.9 & 17.6 \\

\cmidrule(lr){1-1} \cmidrule(lr){2-4} \cmidrule(lr){5-7} \cmidrule(lr){8-10}
Selector & 43.3 & 14.1 & 19.1 & 35.3 & 10.8 & 17.8 & 34.3 & 7.1 & 17.1 \\
\textsc{LED-LC} & 42.0 & 13.3 & 19.2 & 34.9 & 9.9 & 18.1 & 33.9 & 6.6 & 16.8 \\
\textsc{GranuSum}  & \textbf{43.7} & \textbf{14.2} & 20.1 & \textbf{36.0} & \textbf{11.3} & \textbf{18.6} & \textbf{34.8} & \textbf{7.3} & \textbf{17.9} \\
\quad - Ranking & 43.5 & 14.0 & 19.7 & 35.4 & 10.8 & 18.5 & 34.3 & 7.0 & 17.2 \\

\bottomrule
\end{tabular}
\caption{Results of unsupervised abstractive summarization on three datasets.}
\label{tab:unsupervised}
\end{table*}

\paragraph{Automatic Evaluation} As illustrated in Table~\ref{tab:granuduc}, compared to PEGASUS, LED-LC can bring a certain degree of improvement due to the ability to control the length of the output summary. This improvement is not remarkable at fine granularity level. For coarse and medium granularity levels, LED-LC can control the number of output sentences, while PEGASUS does not have a similar capability and it can only generate shorter summaries by truncating the output (to 32 and 64 words), which leads to performance degradation. On the other hand, \textsc{GranuSum} exceeds LED-LC and PEGASUS by a large margin in all the granularity levels. Although \textsc{GranuSum} and LED-LC are trained on the same data, \textsc{GranuSum} increases the R-1 score by 1.78 at coarse granularity level (21.83$\rightarrow$23.61), and the improvement reaches to 4.53 at fine granularity level (30.18$\rightarrow$34.71). With the benefit of event information, our model can generate more relevant and quality summaries, and the advantage is more pronounced in fine-grained summaries. Therefore, \textsc{GranuDUC} is a more suitable system for multi-granularity scenarios than existing controllable summarization models.

\paragraph{Human Evaluation} We also conduct human evaluation to have a more comprehensive understanding of the model output. Six graduate students are involved in this process to score the generated summaries from three different perspectives: fluency, relevance and faithfulness to the source documents. The score range is 1-5, with 1 being the worst and 5 the best. Each sample requires two people to discuss and agree on the scoring. According to the fluency scores in Table~\ref{tab:granuduc}, both LED-LC and \textsc{GranuDUC} can generate coherent sentences, while PEGASUS performs poorly in coarse and medium granularity levels due to truncating the output to a fixed length. From the perspective of relevance and faithfulness, a clear trend is that the more fine-grained the summary, the more relevant it is to the original text and the more likely it is to contain factual errors. Specific to the models, \textsc{GranuSum} generates more relevant and faithful summaries in all granularity scenarios compared to other baselines by exploiting event information.

\subsubsection{Bucket-based Evaluation}
\label{bucket_sec}
In addition to GranuDUC, we seek to utilize existing large-scale datasets for multi-granularity evaluation.
Unlike the previous approach of using a single reference summary to evaluate multiple lengths of summaries~\cite{shapira2018evaluating}, we divide the reference summaries into different buckets based on semantic coverage and then compare the performance of each model in each bucket.
We first design a metric to calculate the granularity score between the source document and the reference summary to categorize the different samples. Because the same events in original text and human-written summary may have different descriptions, we design a granularity score on the basis of BERTScore~\cite{zhang2019bertscore} to perform soft matching due to its ability to measure semantic coverage between two sequences. Specifically, we extract all the events in the source document and the reference summary as two event sequences, and calculate \textbf{\underline{Granularity Score}} as:

\vspace{-2mm}
\begin{align}
\mathrm{Granu}(D, r) = f(Event_D, Event_r),
\end{align}

\noindent where $D$ is the source documents and $r$ represents the reference summary.
$Event_D$ denotes that we extract all events from $D$ by using the approach in Section~\ref{sec:ee}, and  concatenate them into an event sequence.
$f$ means that BERTScore is used to calculate the recall score between two event sequences.
Intuitively, a high recall score of the reference summary to the original text indicates that it has high semantic coverage and thus it is a summary at a high granularity level.
We sort all samples in the test set of Multi-News dataset according to Granularity Score and divide them into three buckets with the same number of samples. The average length of summaries in the three buckets are 198, 214, and 236 words, respectively.

Although PRIMERA is the state-of-the-art model, it does not have the flexibility to change the output in response to different buckets. For LED-LC, we let the model generate 7, 8, and 9 sentences in low, medium, and high buckets, respectively. For our model, we take the top 70\%, 80\%, and 90\% of the events with the higher salience score (see Section~\ref{selector}) in 9 selected sentences as the input for three different buckets. As shown in Table~\ref{tab:bucket}, LED-LC has no significant benefits over PRIMERA, indicating that controlling the output length and ignoring its connection to the original text is not a good solution for the multi-granularity system. In contrast, \textsc{GranuSum} achieves substantial improvements in all buckets compared to powerful baselines. In particular, in buckets with high semantic coverage, our model improves R-1 score by 3.28 compared to PRIMERA. Also, ``- Ranking'' means that we no longer filter out events based on the salience score, which causes a performance drop. It confirms that our selector can indeed exclude irrelevant and redundant events and thus improve the quality of the generated summary.

\subsection{Unsupervised Abstractive Evaluation}
\label{unsupervsied_sec}
The quality of the summary is a key factor for all summarization systems. So in addition to the multi-granularity scenario, we likewise compare \textsc{GranuSum} with conventional unsupervised abstractive summarization models. Table~\ref{tab:unsupervised} provides results on three datasets. The first section includes a simple yet effective approach LEAD, which refers to extracting the first few sentences at the beginning of the text as a summary. It is a strong baseline in the news domain due to the lead bias problem~\cite{see2017get,zhong2019searching}.
The second section lists the strong baselines and the last section contains the results of our models. Selector indicates that we extract several sentences from the source document based on our importance score described in Section~\ref{selector} as the summary. 

Surprisingly, although \textsc{GranuSum} is not specially designed for the conventional unsupervised summarization task, it still beats all the competitors and achieves new state-of-the-art results on most metrics across datasets.
Despite inputting the same hints, PEGASUS-event does not show the ability to exploit event information and even performs worse than PEGASUS.
In contrast, our pre-trained Event-aware Summarizer incorporates event information well into the generated summaries and thus boosts performance.
Furthermore,  \textsc{GranuSum} outperforms Selector, which is a strong extractive baseline, and extractive approaches usually dominate unsupervised summarization tasks.
We think the improvement comes from two reasons:

\noindent 1) In the pre-training stage, important content in the masked sentences is easier to reconstruct due to the redundancy of input texts. Thus, \textsc{GranuSum} learn to filter those unimportant content in inference, generating more concise summaries.

\noindent 2) Event Selector screens out less critical events which should not appear in the summary. 

\noindent Overall, \textsc{GranuSum} improves R-1 score by 1.0 on average compared to the previous best results, indicating that it is sufficient to generate quality summaries besides the multi-granularity ability.

\section{Conclusion}

In this paper, we highlight the importance of multi-granularity summarization systems in catering to user preferences and applying them to real-world scenarios.
To facilitate research in this direction, we propose the first unsupervised multi-granularity summarization framework \textsc{GranuSum} and build a well-established testbed.
Experiments demonstrate the effectiveness of our framework.

\section*{Limitations}

We state the limitations of this paper from the following four aspects:

1) Unlike previous work that uses summary length to approximate granularity, we adopt an event-based definition, which can be extended to be more flexible.
For example, introducing phrases, entities, relationships, etc. as part of the granularity may be a feasible way to further enhance the granularity-aware summarization system.

2) Despite being the first multi-granularity summarization benchmark, GranuDUC can only be used as a test set due to its small size.
Thus, we call for the emergence of customized summarization datasets, which can greatly facilitate the development of customizable summarization models.

3) Specific to the method, we extract events from the source text as hints, which may reduce the abstractness of the generated summaries to some extent.
In pursuit of a more abstractive summary, rephrasing events into different forms may be a viable option, and we leave it as future work.

4) In this paper we focus on three different levels of granularity and take document clusters containing thousands of words as input.
A promising extension could be to input longer text and to add finer levels of granularity, for example, to generate summaries for an entire book (e.g., a novel) at multiple granularities.

\section*{Acknowledgements}
We thank Wen Xiao for providing the output of PRIMERA.
We would also like to thank anonymous reviewers for valuable comments and suggestions.
Research was supported in part by US DARPA KAIROS Program No. FA8750-19-2-1004 and INCAS Program No. HR001121C0165, National Science Foundation IIS-19-56151, IIS-17-41317, and IIS 17-04532, and the Molecule Maker Lab Institute: An AI Research Institutes program supported by NSF under Award No. 2019897, and the Institute for Geospatial Understanding through an Integrative Discovery Environment (I-GUIDE) by NSF under Award No. 2118329. Any opinions, findings, and conclusions or recommendations expressed herein are those of the authors and do not necessarily represent the views, either expressed or implied, of DARPA or the U.S. Government. The views and conclusions contained in this paper are those of the authors and should not be interpreted as representing any funding agencies.

\bibliography{anthology,custom}
\bibliographystyle{acl_natbib}

\appendix

\clearpage
\section{Method}
Here we provide more details about our method part. The workflow of \textsc{GranuSum} and case study are listed in Table~\ref{tab:overall_case}.

\begin{table*}
\centering \footnotesize
\tabcolsep0.15 in
\begin{tabular}{ll}
\toprule
\textbf{Patterns} & \textbf{Examples}  \\  
\midrule
$n_1$-$nsubj$-$v_1$ & Hurricane hit \\ 
$n_1$-$nsubj$-$v_1$-$dobj$-$n_2$ & Hurricane damage buildings \\ 
$n_1$-$nsubj$-$v_1$-$xcomp$-$a$ & People feel scared \\
$n_1$-$nsubj$-$v_1$-$xcomp$-$v_2$-$dobj$-$n_2$ & Police want to save people \\
$n_1$-$nsubjpass$-$v_1$ & Residents are injured \\
\bottomrule
\end{tabular}
\caption{Five typical patterns and corresponding examples when we extract events (76 patterns in total). Here ‘v’ is a verb, ‘n’ stands for a noun, and ‘a’ denotes an adjective. All verbs remain in their original form. ‘nsubj’, ‘dobj’, ‘xcomp’, and ‘nsubjpass’ are syntactic relations. }
\label{tab:pattern}
\end{table*}

\begin{table*}[t]
\center \footnotesize
\begin{tabular}{p{15cm}}
\toprule
\textbf{Step 1: Select Important Sentences based on Relevance and Redundancy Score, and Extract Events} \\

    \quad\tabitem \colorY{Malone} was part of the 76ers' 1983 NBA championship team, and the \colorY{club said} he will forever \colorY{be remembered} as a genuine icon and pillar of the most storied era in the history of Philadelphia 76ers basketball. $\longrightarrow$ \colorY{club say} | \colorY{Malone be remember}  \\
    \quad\tabitem In the initial meeting in New York, \colorY{Cunningham pulled Malone} aside and \colorY{let him know} his expectations of the player who had \colorY{won MVP} honors in Houston the previous season by \colorY{averaging 31.1 points and 14.7 rebounds}. $\longrightarrow$ \colorY{Cunningham pull Malone} | \colorY{let hime know} | \colorY{win MVP} | \colorY{average 31.1 points and 14.7 rebounds} \\
    \quad\tabitem In his first season with the Sixers, \colorY{Malone won MVP} awards by \colorY{averaging 24.5 points and 15.3 rebounds} during the regular season in which the \colorY{team compiled a 65-17 record}. $\longrightarrow$ \colorY{Malone win MVP} | \colorY{average 24.5 points and 15.3 rebounds} | \colorY{team compile a 65-17 record} \\
    \quad\tabitem \colorY{Moses Malone}, a three-time NBA MVP and one of basketball's most ferocious rebounders, \colorY{died} Sunday, the Philadelphia \colorY{76ers said}. $\longrightarrow$ \colorY{Moses Malone die} | \colorY{76ers say} \\

\midrule
\midrule

\textbf{Step 2: Obtain a Candidate Set by Combining the Above Events} \\

\quad\tabitem Original Candidate Events: \colorG{club say} | Malone be remember | Cunningham pull Malone | let him know | \colorB{win MVP} | \colorR{average 31.1 points and 14.7 rebounds} | \colorB{Malone win MVP} | \colorR{average 24.5 points and 15.3 rebounds} | team compile a 65-17 record | Moses Malone die | \colorG{76ers say} \\

\midrule
\midrule

\textbf{Step 3: Event Ranking and Filtering (Event Selector)} \\

\quad\tabitem Ranked Candidate Events: \colorB{Malone win MVP} | Moses Malone die | Malone be remember | team compile a 65-17 record | Cunningham pull Malone | \colorR{average 31.1 points and 14.7 rebounds} | \colorG{76ers say} | let him know \\
\midrule
\midrule
\textbf{Step 4: Multi-Granularity Summary Generation (Event-based Summarizer)} \\

\quad\tabitem Coarse Granularity Level\\
\quad\tabsubitem Input: Malone win MVP | Moses Malone die $\langle$seg$\rangle$ $\langle$mask$\rangle$ Source News \\
\quad\tabsubitem Generated Summary: \underline{Moses Malone, a three-time NBA MVP} and one of basketball's most ferocious rebounders, \underline{died} on Sunday. \\

\quad\tabitem Fine Granularity Level\\
\quad\tabsubitem Input: Malone win MVP | Moses Malone die | Malone be remember | team compile a 65-17 record $\langle$seg$\rangle$ $\langle$mask$\rangle$ Source News \\
\quad\tabsubitem Generated Summary: \underline{Moses Malone, a three-time NBA MVP} and one of basketball's most ferocious rebounders, \underline{died} on Sunday.
He helped the team compile a 65-17 record in the first season. These achievements \underline{make him be remembered as a genuine icon and pillar in the history of 76ers basketball}. \\

\midrule
\midrule

\textbf{Summary Generated by PEGASUS}

\tabsubitem \underline{Moses Malone, a three-time NBA MVP} and one of basketball's most ferocious rebounders, \underline{died} Sunday, the Philadelphia 76ers said. The 76ers issued a statement that said Malone had died. Malone was inducted into the Naismith Memorial Basketball Hall of Fame in 2001 and attended the induction ceremonies for the year's class in Springfield, Massachusetts this weekend. \\

\midrule
\midrule

\textbf{Reference Summary}

\tabsubitem \underline{Three-time NBA MVP and Philadelphia 76ers legend Moses Malone}, who with Julius Erving in 1983 brought the City of Brotherly Love its first championship since 1967, \underline{has died} at the age of 60, reports the Inquirer. \underline{Moses holds a special place in our hearts and will forever be remembered as a genuine icon and pillar} of the most storied era in the history of Philadelphia 76ers basketball. \\

\bottomrule

\end{tabular}
\caption{Workflow of \textsc{GranuSum} and case study. The colored text in Step 1 indicates the location of the extracted event in the original sentence. Events of the same color in Step 2 are redundant. Underlined text in Step 4 represents the overlap with the reference summary. Notably, we pre-train an Event-based Summarizer before Step 1.}
\label{tab:overall_case}
\end{table*}

\subsection{Event Extraction}
\label{sec:appendix_ee}
Specifically, given a sentence $s$, we use a dependency parser to obtain its dependency parse tree and select all non-auxiliary verbs as centric tokens.
Then, along the syntactic relationships between the selected verbs and other tokens, we extract the longest phrase that matches the designed patterns as events.
As illustrated in Table~\ref{tab:pattern}, the most frequent pattern is $n_1$-nsubj-$v_1$, such as \textit{Hurricane hit}. Another common pattern is $n_1$-nsubj-$v_1$-dobj-$n_2$, like \textit{Hurricane damage buildings}.
Here ``nsubj'' denotes an active relationship between nouns and verbs, while ``nsubjpass'' in another example represents a passive relationship between them. More detailed examples can be found in Table~\ref{tab:overall_case}, we extract events from four selected sentences, and the colored text shows the locations of the events in the original document.

\subsection{Event-based Summarizer Pre-training}
\label{sec:appendix_summarizer}
We further explain the four steps of Event-based Summarizer pre-training with the help of the following example.
For a paragraph of news as \textit{``Honduras braced for potential catastrophe Tuesday. Hurricane Mitch roared through the Caribbean, churning up high waves and intense rain that sent coastal residents scurrying for safer ground. President declared a state of maximum alert and the Honduran military sent planes to pluck residents from their homes on islands near the coast''}, we

1) first randomly select a sentence: \textit{``Hurricane Mitch roared through the Caribbean, churning up high waves and intense rain that sent coastal residents scurrying for safer ground''},

2) extract events in it such as \textit{Mitch roar}, \textit{Mitch churn up wave and rain}, \textit{send} and \textit{resident scurry},

3) then mask this sentence in the original paragraph, and finally

4) use extracted events and masked text as the input and regard the selected sentence as the target as follows:

\begin{itemize}
    \item Input: Mitch roar | Mitch churn up wave and rain | send | resident scurry $\langle$seg$\rangle$ Honduras braced for potential catastrophe Tuesday. $\langle$mask$\rangle$ President declared a state of maximum alert and the Honduran military sent planes to pluck residents from their homes on islands near the coast.
    \item Target: Hurricane Mitch roared through the Caribbean, churning up high waves and intense rain that sent coastal residents scurrying for safer ground.
\end{itemize}

In our experiments, we randomly mask 1 to $n$ sentences from a document, which leads to $n$ samples to pre-train our Summarizer.
Here we set $n$ to the smaller of a constant number 10 and one-third of the number of sentences in the document.

\subsection{Event Selector}
\label{sec:appendix_selector}
We use the example in Table ~\ref{tab:overall_case} to further explain the flow of the Event Selector.
When we obtain candidate events from selected sentences, there are still different types of issues in the candidate set. Some generic and uninformative events, such as ``\textit{club say}'' and ``\textit{let him know}'', should have a lower priority for a summary. Although we introduce sentence-level redundancy score in the pruning step, as a finer-grained unit, events still suffer from redundancy problem (see events in Table~\ref{tab:overall_case} with the same color), e.g., both ``win MVP'', ``Malone win MVP'' and ``average 31.1 points and 14.7 rebounds'', ``average 24.5 points and 15.3 rebounds'' appear in the candidate set. However, after the events ranking and filter using our Event Selector, all of these issues are alleviated. In this case, our Selector regards ``Malone win MVP'', ``Moses Malone die'' and ``Malone be remember'' as the three most salient events, which is consistent with the original news. In addition, uninformative events (``\textit{club say}'' and ``\textit{let him know}'') are ranked at the end of the candidate sets, and duplicate events (``win MVP'' and ``average 24.5 points and 15.3 rebounds'') are filtered out due to the lowest salience score. In general, the reasonable ranking of candidate events by 
the Selector plays a crucial role in improving the quality of subsequent multi-granularity summaries.

\subsection{Multi-Granularity Summary Generation}
\label{sec:appendix_generation}
We can see from Table~\ref{tab:overall_case}, to obtain the most condensed summary, the two most important events (``Malone win MVP'' and ``Moses Malone die'') and the original news are fed to the model. Then, the pre-trained Summarizer can be aware of event-based cues and generate the corresponding sentence: ``Moses Malone, a three-time NBA MVP and one of basketball's most ferocious rebounders, died on Sunday''. As more events are input, our Summarizer also has the ability to adjust the order of the narrative to make the content more logical. In the summary of granularity level 2, the order in the prompt is ``Malone be remember'' then ``team compile a 65-17 record'', but the model first output "He helped the team compile a 65-17 record in the first season" and then "These achievements make him be remembered as a genuine icon and pillar in the history of 76ers basketball" to make the whole summary more coherent and intuitive.
Compared to sentences selected from the source documents (see Step 1 in Table~\ref{tab:overall_case}), the summary generated by GranuSum omits unimportant details and paraphrases to make it more concise.
Abstractive models without guidance signals, such as PEGASUS, tend to generate some repetitive sentences (the first two sentences), and generate several less relevant sentences without capturing important events.
In contrast, \textsc{GranuSum} can output summaries that are more relevant and faithful to the original text.

\section{Examples for GranuDUC}

We provide two annotation examples for our proposed GranuDUC benchmark in Table~\ref{tab:annotation}.

\begin{table*}[t]
\center \footnotesize
\begin{tabular}{p{15cm}}
\toprule

\textbf{Sample 1: News about the Civil Suit against Microsoft} \\

    \quad\tabitem \textbf{Summary of Coarse Granularity Level}: The Justice Department filed a civil suit against Microsoft to change its pattern of anti-competitive conduct on browser software.  \\
    \quad\tabitem \textbf{Summary of Medium Granularity Level}: Business rivals have filed an anti-trust suit against Microsoft to break Microsoft Corp.'s monopoly on computer operating systems. The suit began with a Microsoft vs Netscape battle. The Government is examining Microsoft's financial records and painting a dark image of its Chairman Bill Gates. An unpublished book may be crucial to the trial. \\
    \quad\tabitem\textbf{Summary of Fine Granularity Level}: The Justice Department filed a suit against Microsoft for violation of the Sherman Act to change its anti-competitive conduct. The heart of the suit is the Internet browser battle between Microsoft and Netscape. Microsoft, it is argued, has told computer manufacturers that if they want Windows, they must forgo Netscape. Netscape complaint over browsers was central to the case, which grew to include Intel, IBM, Sun, Apple, AOL, and Intuit. The battle now extends far beyond that aiming at Microsoft's overall aggressive anti-competitive conduct. Microsoft's chairman, Bill Gates, usually seen as a visionary is portrayed in much darker tones in the trial. Microsoft was ordered to let Justice examine its records and sought a trial delay. An unpublished book provided evidence, which can be crucial to the trial. \\

\midrule

\textbf{Sample 2: News about the Health Condition of the Russian President} \\

    \quad\tabitem \textbf{Summary of Coarse Granularity Level}: Russia President Boris Yeltsin's worsening heath condition caused great concern to the Russian leadership.  \\
    \quad\tabitem \textbf{Summary of Medium Granularity Level}: During Russia President Boris Yeltsin's seven years in power, illness has often sidelined him. He recently cut short a trip to Central Asia because of a respiratory infection and he later canceled two out-of-country summits. Russia's leaders are calling for his resignation and question his legal right to seek reelection. \\
    \quad\tabitem\textbf{Summary of Fine Granularity Level}: Russia President Boris Yeltsin had a heart attack in 1996, followed by multiple bypass surgery. The cause of minor burns on his hand were not disclosed. On a trip to Uzbekistan he walked stiffly, stumbled, rambled and seemed confused. Ceremonies were canceled and the trip ended a day early. Yeltsin refuses to admit he is seriously ill and his condition is kept secret. He was treated with antibiotics and ordered to bed but went to the office anyway. Many Russians suspect he is sicker, question his ability to do his job, and want him to resign. The court was to judge on whether he could serve a third term, but he already has said he will not run. \\

\bottomrule

\end{tabular}
\caption{Annotation of two samples in GranuDUC.}
\label{tab:annotation}
\end{table*}

\end{document}